\documentclass[journal]{IEEEtran}
\usepackage{amsmath,amsfonts}
\usepackage{algorithmic}
\usepackage{algorithm}
\usepackage{array}
\usepackage[caption=false,font=normalsize,labelfont=sf,textfont=sf]{subfig}
\usepackage{textcomp}
\usepackage{stfloats}
\usepackage{url}
\usepackage{verbatim}
\usepackage{graphicx}
\usepackage{cite}
\begin{document}

\title{CORE: Toward Ubiquitous 6G Intelligence Through Collaborative Orchestration of Large Language Model Agents Over Hierarchical Edge}

\author{Zitong~Yu, Boquan~Sun, Yang~Li, Zheyan~Qu, Xing~Zhang
    \thanks{Zitong Yu, Boquan Sun, Yang Li, Zheyan Qu, Xing Zhang (corresponding author) are with the Beijing University of Posts and Telecommunications, Beijing, China. E-mail: hszhang@bupt.edu.cn.This work was supported by the National Science Foundation of China under Grant 62271062.}
}

\markboth{Magazine of \LaTeX\ Class Files,~Vol.~14, No.~8, July~2025}
{Shell \MakeLowercase{\textit{et al.}}: CORE: Toward Ubiquitous 6G Intelligence Through Collaborative Orchestration of Large Language Model Agents Over Hierarchical Edge}

\maketitle

\begin{abstract}
Rapid advancements in sixth-generation (6G) networks and large language models (LLMs) have paved the way for ubiquitous intelligence, wherein seamless connectivity and distributed artificial intelligence (AI) have revolutionized various aspects of our lives.  However, realizing this vision faces significant challenges owing to the fragmented and heterogeneous computing resources across hierarchical networks, which are insufficient for individual LLM agents to perform complex reasoning tasks.  To address this issue, we propose Collaborative Orchestration Role at Edge (CORE), an innovative framework that employs a collaborative learning system in which multiple LLMs, each assigned a distinct functional role, are distributed across mobile devices and tiered edge servers. The system integrates three optimization modules, encompassing real-time perception, dynamic role orchestration, and pipeline-parallel execution, to facilitate efficient and rapid collaboration among distributed agents. Furthermore, we introduce a novel role affinity scheduling algorithm for dynamically orchestrating LLM role assignments across the hierarchical edge infrastructure, intelligently matching computational demands with available dispersed resources. Finally, comprehensive case studies and performance evaluations across various 6G application scenarios demonstrated the efficacy of CORE, revealing significant enhancements in the system efficiency and task completion rates. Building on these promising outcomes, we further validated the practical applicability of CORE by deploying it on a real-world edge-computing platform, that exhibits robust performance in operational environments.
\end{abstract}

\begin{IEEEkeywords}
6G networks, Large language models (LLMs), AI agents, Hierarchical edge computing, Ubiquitous intelligence.
\end{IEEEkeywords}

\section{Introduction}

\IEEEPARstart{T}{he} convergence of sixth-generation (6G) wireless networks and large language models (LLMs) heralds a transformative era of ubiquitous intelligence, reshaping interactions in domains such as smart cities, healthcare, and industrial automation~\cite{Xu2024}. Unlike their predecessors, 6G networks offer ultralow latency, extensive connectivity, and unprecedented data rates, thereby enabling LLM-powered AI agents at the network edge. These agents, transitioning from centralized cloud-based AI to distributed, edge-native entities, leverage pervasive computational resources to deliver context-aware, real-time services. Ubiquitous intelligence envisions a future in which AI agents, empowered by LLMs, operate autonomously and collaboratively to provide personalized, efficient, and scalable solutions. This paradigm shift enhances user experience and unlocks the full potential of 6G ecosystems~\cite{Cui2025}, which is promising for redefining human-machine interactions with unprecedented levels of automation, personalization, and efficiency.
Current approaches to deploying LLMs in 6G networks often rely on centralized cloud systems or isolated edge devices, but both face limitations. Centralized solutions incur high latency owing to the distance between users and data centers, making them unsuitable for time-critical applications, such as autonomous driving or healthcare diagnostics. Conversely, edge devices, constrained by limited computational power, struggle with complex tasks and can become overburdened, degrading performance and increasing energy use. These shortcomings highlight the need for a new approach for effectively harness 6G’s distributed resources effectively.

The fragmented and heterogeneous nature of computing resources across 6G networks presents significant challenges. Computational power spans resource-constrained mobile devices, edge servers, and cloud infrastructure, each with varying capabilities and constraints~\cite{Wang2025}. Individual LLM agents, tasked with complex reasoning activities such as multi-modal perception, dynamic decision-making, or natural language understanding often lack the resources to operate efficiently in isolation. In time-sensitive applications such as autonomous vehicles immersive digital twins, or real-time healthcare diagnostics, low-latency collaboration among agents is critical. However, the current landscape, characterized by isolated and overburdened resources, hinders effective agent coordination and limits the scalability and reliability of AI-driven services~\cite{Zheng2023}. Addressing these challenges is technically imperative to realize the transformative potential of ubiquitous intelligence in 6G networks.

To overcome these obstacles, we propose a Collaborative Orchestration Role at Edge (CORE), an innovative framework designed to enable collaborative execution of user-agent interactive tasks in 6G networks. CORE leverages a collaborative learning system in which multiple LLMs, each assigned a distinct functional role, are distributed across mobile devices and tiered-edge servers. It integrates real-time perception, dynamic role orchestration, and pipeline-parallel execution to enable efficient agent collaboration. The effectiveness of CORE is demonstrated through its deployment on a real-world edge computing platform for industrial automation, where it powers a multi-agent system for real-time anomaly detection.

Specifically, CORE decomposes complex tasks into manageable subtasks, assigns them to specialized LLM roles, and orchestrates their execution across the network hierarchy. Network-aware orchestration optimizes resource allocation by considering latency, bandwidth, and device capabilities, whereas pipeline-parallel execution enhances throughput through concurrent processing across tiers. Furthermore, we present a novel role-affinity scheduling algorithm that dynamically assigns LLM roles to available resources based on the task requirements, network conditions, and device capabilities. This algorithm improves system adaptability by minimizing execution delays while ensuring task completion quality, even in dynamic 6G environments characterized by fluctuating workloads and network variability. By harmonizing these components, CORE enables seamless user-agent interactions and supports the deployment of robust AI-driven services across diverse 6G scenarios.

Overall, the primary contributions of this article are summarized as follows:
\begin{itemize}
    \item We introduced CORE to facilitate more efficient collaboration among distributed multi role agents within 6G networks and validated its effectiveness through deployment on an edge-computing platform for industrial automation.
    \item We present an optimization framework that effectively integrates real-time perception, dynamic orchestration, and pipeline-parallel execution to enhance the task completion rate and system efficiency of multi-agent systems in diverse 6G scenarios.
    \item We developed a role-affinity scheduling algorithm designed to optimize role assignments among distributed agents, thereby effectively utilizing hierarchical edge resources within 6G networks.
\end{itemize}

\section{6G-Enabled LLM Agents: Advancing Ubiquitous Intelligence}
\label{sec:6g_llm}

The integration of sixth-generation (6G) wireless networks with Large Language Model (LLM) agents represents a significant advancement in the field of intelligent systems. This section examines the opportunities and challenges associated with this convergence, highlighting the technological enablers of 6G and the crucial role of LLM agents in the development of smart networks.

\subsection{6G Technologies Empowering LLM Agents}

\subsubsection{Opportunities of 6G for Advancing LLM Agents}

The advanced capabilities of 6G networks substantially enhance LLM agents through significant advancements in data capacity, latency, connectivity, and AI-native designs.  With a per-user bandwidth of 1 Tbps, 6G facilitates the real-time transmission of multi-modal data, such as 4K/8K video and 3D point clouds, thereby enabling LLM agents to execute sophisticated multi-modal fusion and process complex data streams \cite{Hang2024}. The ultralow air interface latency of 6G, which is less than 1 ms, in conjunction with edge computing, supports rapid "perception-decision-execution" cycles\cite{Stoica2019}.  Furthermore, its extensive connectivity, which accommodates up to $10^6$ devices per square kilometer, allows LLM agents to aggregate real-time data from expansive smart city sensor networks, thereby improving situational awareness and decision accuracy \cite{Xu2024}. Notably, 6G's integrated sensing and communication (ISAC) facilitates synesthesia-like multi-modal fusion, which is vital for applications in autonomous driving and healthcare \cite{Cui2025}. Its AI-native architecture, which utilizes network slicing and Network Function Virtualization(NFV), supports self-optimizing networks, where LLM agents can predict traffic, dynamically allocate resources, and optimize efficiency, rendering 6G indispensable for enhancing the responsiveness and scalability of LLM agents \cite{Ahammed2023}.

\begin{figure}[t]
    \centering
    \includegraphics[width=0.48\textwidth]{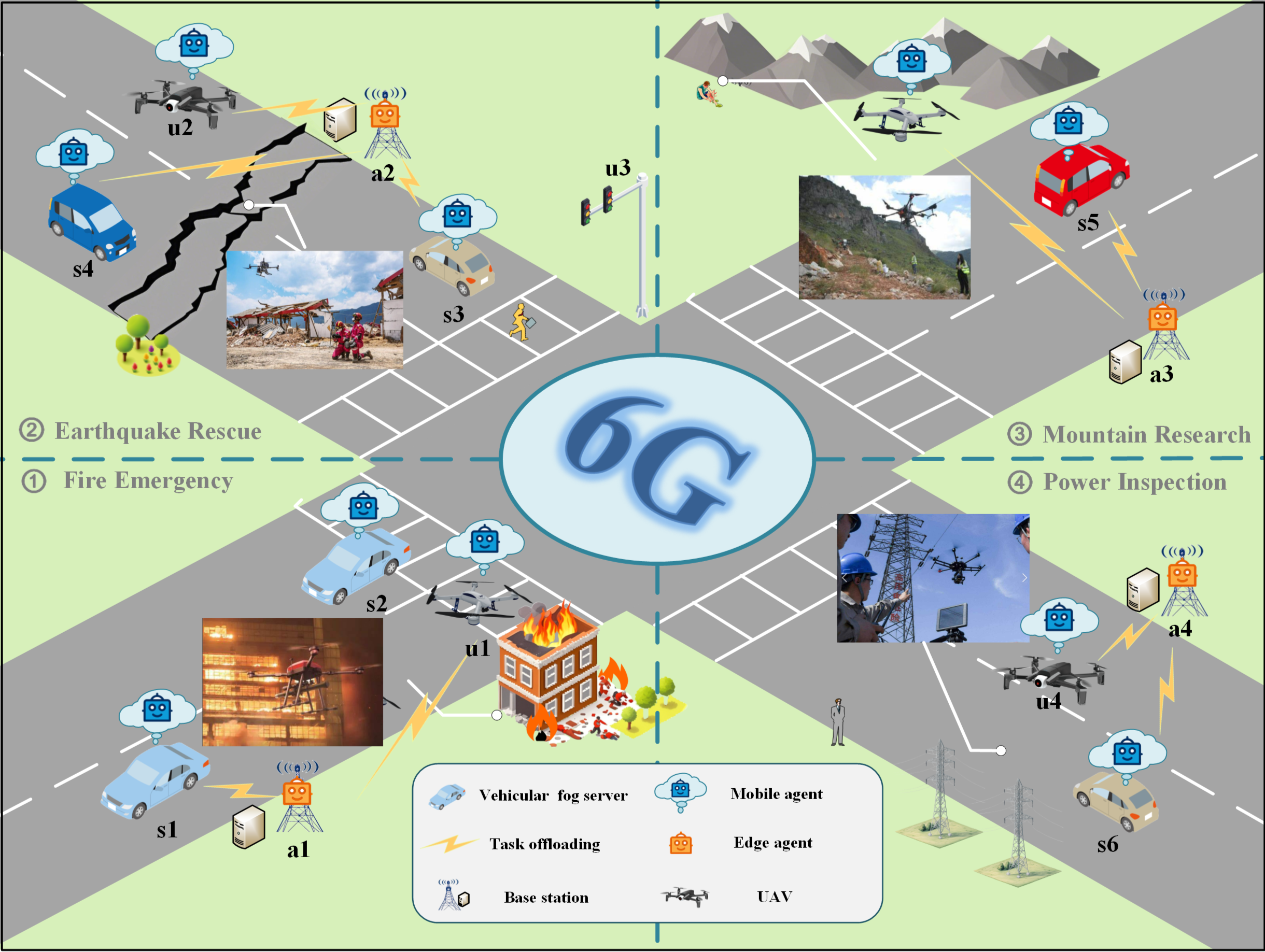}
    \caption{Multi AI Agents in Ubiquitous 6G Intelligence.}
    \label{fig:ubiquitous_ai}
\end{figure}

\subsubsection{Ubiquitous 6G Intelligence}

In the 6G era, the concept of ubiquitous intelligence is fundamentally based on the integration of LLM agents within a hierarchical edge-cloud architecture designed to optimally balance performance, efficiency, and latency. This tripartite framework consists of the device layer, which is responsible for managing highly latency-sensitive local tasks, such as on-device video analysis; the edge layer, which performs intermediate processing tasks, including real-time language understanding for smart homes, thereby significantly reducing the transmission of raw data to the cloud~\cite{Zhang2025}; and the cloud layer, which supports complex cross-domain model training and global optimization. This proximate layered design minimizes latency while ensuring scalability and robustness. Leveraging this architecture and 6G's ultra-low latency and high bandwidth, LLM agents drive transformative applications across diverse sectors, as illustrated in Figure~\ref{fig:ubiquitous_ai}. In smart cities, traffic signals are optimized, energy grids are managed, and anomalies are detected. For space-air-ground integrated networks, LLMs synthesize data from satellites, drones, and terrestrial sensors to enhance precision agriculture and coordinate disaster responses. In industrial automation, they predict equipment failures and optimize production lines, whereas in smart healthcare, they enable remote monitoring, medical image analysis, and telemedicine \cite{Zheng2023, Wang2025}. Thus, the synergy between the 6G hierarchical architecture and LLM agents constitutes the core technological pathway towards truly pervasive and adaptive intelligent systems.

\subsubsection{Coexistence and Tension: LLMs in the URLLC Context}

Sixth-generation (6G) networks necessitate stringent ultra-reliable low-latency communication (URLLC) requirements, which demand sub-millisecond latency and near-perfect reliability for real-time control applications. However, even the most advanced LLMs operating on edge devices typically require hundreds of milliseconds for inference, rendering them fundamentally incompatible with these constraints. Consequently, a latency-aware task-allocation strategy is imperative. LLMs should be allocated to delay-tolerant high-value tasks, such as predictive maintenance scheduling, long-term urban planning, and offline data analysis, where their superior reasoning and generalization capabilities offer distinct advantages. In these contexts, techniques such as pipeline parallelism can further enhance the throughput without compromising the system objectives. Conversely, safety-critical tasks that necessitate strict URLLC compliance, including collision avoidance, remote surgery, and industrial safety, require predictable sub-millisecond responses. These functions must rely on simple rule-based systems or highly optimized lightweight models, rather than LLMs. Thus, the effective integration of LLMs in 6G depends on hybrid architectures: LLMs function at a higher hierarchical level to evaluate the system state and issue high-level directives, whereas URLLC-compliant agents execute immediate perception and actuation.

\subsection{LLM Agents: The Neurons of the Smart Network}

\subsubsection{Roles of LLM Agents in 6G Networks}

LLM agents play a crucial role in harnessing the capabilities of 6G networks and act as intelligent facilitators across various functional domains. As perception orchestrators, LLMs synthesize multi-modal data, including LiDAR point clouds and camera feeds, into coherent semantic representations. In the realm of traffic monitoring, the integration of CLIP-based models has markedly enhanced object detection accuracy, achieving improvements of 10–20 pp in certain scenarios~\cite{Ahammed2023}.As dynamic reasoning engines, LLMs respond instantaneously to network behaviour to mitigate the resource limitations. For instance, during virtual reality streaming congestion, LLMs employ reinforcement learning techniques, such as Proximal Policy Optimization (PPO) and Dynamic Policy Optimization (DPO), to dynamically enhance model precision and resource allocation. This approach reduces bandwidth consumption by 20--50\% while maintaining high frame rates (e.g., 60 FPS)~\cite{Wang2025b}. In addition, LLMs serve as network intelligence coordinators that manage tasks across distributed edge nodes. In wildfire response scenarios, LLM agents coordinate drone fleets to disseminate satellite imagery analyses, thereby significantly reducing processing time. In the context of facilitating federated learning, LLMs integrate hospital AI models with differential privacy, thereby enhancing cancer detection performance while significantly mitigating privacy risks through the application of differential privacy~\cite{Jiang2023,Wang2025}.By leveraging the ultralow latency and high bandwidth of 6G, LLM agents enable robust, adaptive, and privacy-preserving operation.

\subsubsection{Deployment Challenges}

Integrating LLM agents into 6G networks holds transformative potential; however, it also presents significant challenges. The substantial computational and memory requirements of LLMs impose a considerable strain on edge infrastructure. Numerous LLM-based edge tasks fail because of substantial memory demands that frequently exceed the limited resources available on edge servers. Although quantization alleviates memory pressure, it often compromises accuracy, as evidenced by potential increases in error rates in sensitive tasks, such as medical diagnosis~\cite{Wang2023,Wang2025b}. Furthermore, distributed execution exacerbates LLM error propagation and inconsistency. In medical diagnostics, discrepancies among edge nodes can increase error rates, particularly in safety-critical applications such as medical diagnostics, thereby threatening safety-critical reliability. Additionally, interoperability and synchronization remain problematic; a notable portion of cross-edge tasks may fail because of timing mismatches, network delays, and heterogeneous frameworks, impairing real-time applications such as traffic management, whereas heterogeneous APIs and frameworks further impede seamless collaboration and scalability. Addressing these challenges necessitates concerted advances in model compression, energy-efficient inference, robust distributed consensus mechanisms, and standardized edge-AI interfaces. Only through such innovations can LLM-enabled 6G systems achieve the requisite balance of performance, reliability, and efficiency.

\begin{figure*}[t]
    \centering
    \includegraphics[width=0.88\textwidth]{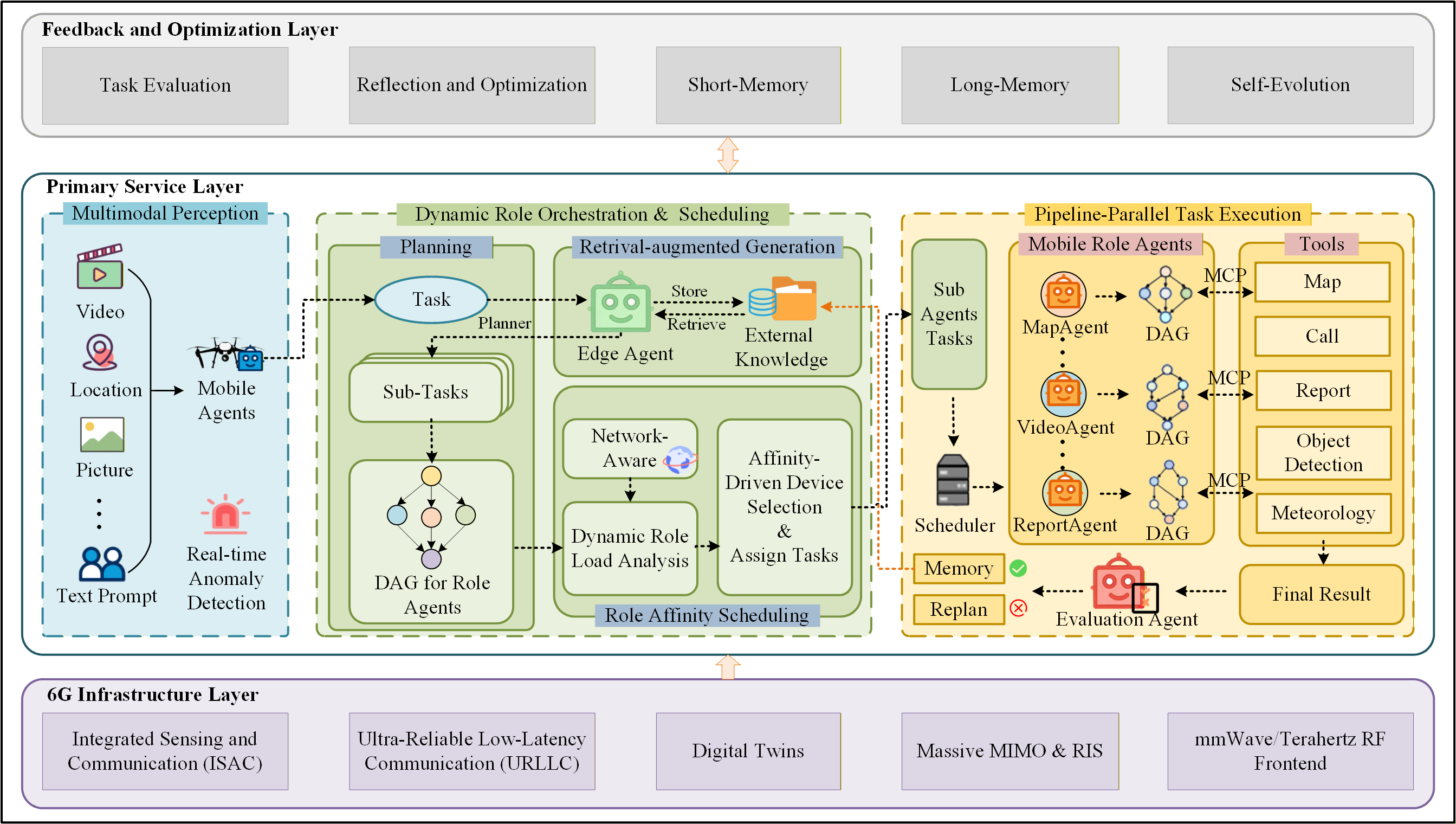}
    \caption{Overview of the CORE framework's three-layer architecture. The top layer is the Feedback and Optimization Layer, the middle layer is the Primary Service Layer where the main CORE modules reside, and the bottom layer is the 6G Infrastructure Layer providing underlying support.}
    \label{fig:core_arch}
\end{figure*}

\begin{figure*}[t]
    \centering
    \includegraphics[width=0.88\textwidth]{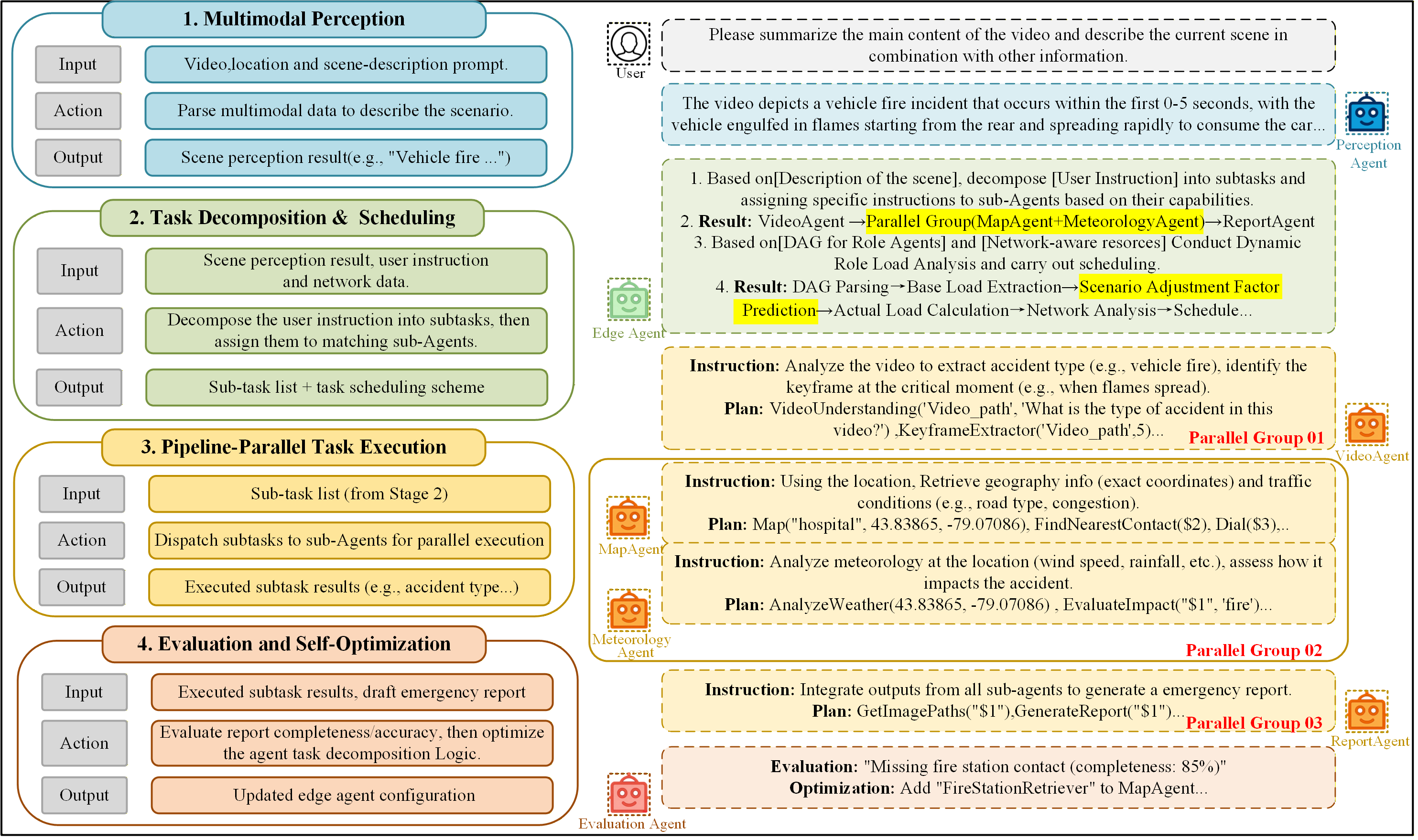}
    \caption{Case study demonstrating dynamic role orchestration in the CORE framework for drone-based emergency rescue. A vehicle fire incident is processed through multi-modal perception, task decomposition, parallel sub-task execution by specialized agents via terminal-edge collaboration, and final evaluation.}
    \label{fig:algo_flow}
\end{figure*}

\section{CORE Framework Architecture}

\label{sec:core}

The Collaborative Orchestration Role at Edge (CORE) framework constitutes an innovative methodology for orchestrating Large Language Model (LLM) agents within sixth-generation (6G) wireless networks, thereby facilitating the efficient and scalable execution of interactive tasks between users and agents. By addressing the challenges posed by resource heterogeneity and the demands for ultra-low latency, CORE enables the integration of ubiquitous intelligence across a wide range of applications, including smart cities and healthcare. This section outlines the architecture of CORE, detailing its system overview, collaborative learning system, multi-tiered optimization framework, and role affinity scheduling algorithm, as illustrated in the framework architecture (Figure~\ref{fig:core_arch}).

\subsection{System Overview}

The CORE framework is structured into three primary layers: the Feedback and Optimization Layer, the Primary Service Layer, and 6G Infrastructure Layer, which are designed to optimize resource utilization and ensure adaptability in dynamic 6G environments. 

The Feedback and Optimization Layer drives continuous improvement through components such as task evaluation, reflection and optimization, short-term memory, long-term memory, and self-evolution. Short-term memory retains recent task contexts and immediate outcomes within a sliding window, whereas long-term memory persistently stores validated agent strategies, historical performance patterns and lessons learned. Both memories are exposed as invocable tools through standardized API interfaces, enabling the reflection module to query past experiences, reinforce successful orchestration decisions, and avoid previously identified erroneous scheduling and role assignments. Task evaluation assesses performance metrics, whereas Reflection and Optimization leverage these memory tools to iteratively refine policies and support self-evolution.

 The Primary Service Layer comprises multi-modal perception, dynamic role orchestration and scheduling, retrieval-augmented generation and pipeline-parallel task execution. multi-modal Perception processes diverse data inputs (e.g., video, location, and text) via Mobile Agents to perform real-time anomaly detection. Dynamic Role Orchestration and Scheduling, enhanced by external knowledge, plans, and assigns tasks by coordinating mobile role agents (e.g., MapAgent, VideoAgent) through edge agents and role affinity scheduling, thereby optimizing the utilization of fragmented computational resources and enhancing task completion rates. Pipeline-parallel task execution leverages directed acyclic graphs (DAGs) and the Model Context Protocol (MCP) to coordinate the use of external API tools to achieve this parallelism. 

The 6G Infrastructure Layer underpins CORE with technologies such as Integrated Sensing and Communication (ISAC), Ultra-Reliable Low-Latency Communication (URLLC), Digital Twins, Edge Intelligence, and the Cloud-Edge-Device Continuum, ensuring robust communication and computation \cite{Loven2025}.

\subsection{Collaborative Learning System}

The collaborative learning system of CORE distributes LLMs across mobile devices and tiered-edge servers and assigns distinct functional roles to leverage their specialized capabilities. This system ensures efficient resource utilization and seamless inter-agent coordination. LLMs are strategically deployed on mobile devices for lightweight tasks, such as local feature extraction, and on edge servers for computationally intensive operations, such as semantic analysis, which minimizes latency by processing data near its source. Each LLM is assigned a role based on task requirements and device capabilities, such as perception, decision making, or communication.  

Inter-agent coordination is facilitated by the Model Context Protocol (MCP), a domain-agnostic context management and transfer mechanism that ensures a consistent contextual understanding across heterogeneous technological domains. The MCP achieves cross-domain interoperability through standardized semantic abstractions, lightweight protocol adapters, and ontology-based payload mapping, enabling seamless context exchange between, for instance, ISAC-generated sensing metadata in the infrastructure layer and application-specific API states in the service layer. Directed acyclic graphs (DAGs) complement the MCP by explicitly modeling task dependencies and execution order, allowing agents in disparate domains, ranging from satellite-based MapAgents to ground-based VideoAgents, to maintain unified workflow awareness and achieve synergistic task completion\cite{Jiang2023}.

\subsection{Multi-tiered Optimization Framework}

The multi-tiered optimization framework integrates real-time multi-modal perception, dynamic role orchestration, and pipeline-parallel execution to address the resource heterogeneity and latency challenges in 6G networks.

Real-time multi-modal perception employs Mobile Agents to process diverse data streams, including videos, locations, and text, using techniques such as object detection and anomaly identification. Dynamic role orchestration assigns roles based on task requirements, network conditions, and device capabilities. Network-Aware Dynamic Role Load Analysis evaluates bandwidth and latency, whereas Role Affinity Scheduling optimizes task assignments. In critical scenarios, such as emergency management, where a digital twin (DT) may be invoked to test and refine scheduling decisions, the framework employs a predictive strategy to guarantee ultra-low latency. Instead of conducting computationally intensive, online DT simulations during each decision cycle, CORE leverages historical data and offline DT simulations to pre-train a lightweight “decision evaluation model” and to generate a “scheduling policy lookup table.” During online operation, the scheduling algorithm directly queries this model or table, achieving the benefits of a DT-based assessment with minimal computational overhead and thereby avoiding the high latency of real-time simulations. This approach ensures that even DT-enhanced orchestration complies with the stringent latency requirements of time-sensitive applications.

The pipeline-parallel execution decomposes tasks into subtasks that are executed concurrently across agents. DAGs ensure proper sequencing, and MCP provides contextual synchronization, thereby enhancing the throughput for data-intensive tasks, such as real-time video processing in smart cities. This framework ensures a high performance and low latency, which are critical for time-sensitive 6G applications.

\subsection{Role Affinity Scheduling Algorithm}

The Role Affinity Scheduling algorithm optimizes task assignments to LLM agents, adapting to the dynamic conditions of 6G environments. Its design leverages an affinity-driven approach, wherein agents are selected based on their suitability for specific tasks, which are calculated using task requirements, network conditions, and device capabilities.

The algorithm considers task complexity, data volume, and real-time constraints, along with network parameters such as bandwidth, latency, and reliability. Device capabilities, including processing power, memory, and energy efficiency, were also evaluated to prevent overload or battery depletion. For instance, a task requiring high-speed processing is assigned to a powerful edge server, whereas an energy-constrained task targets a low-power device.

Adaptive mechanisms enable the algorithm to respond to fluctuating workloads and network variability. Continuous monitoring of network states and agent performance allows for the dynamic reassignment of tasks. If an agent becomes overloaded or experiences latency spikes, the algorithm redistributes the tasks to maintain performance. This adaptability ensures efficient resource utilization and reliable task execution across the diverse 6G scenarios.

\section{Case Study}

\begin{figure}[t]
    \centering
    \includegraphics[width=0.48\textwidth]{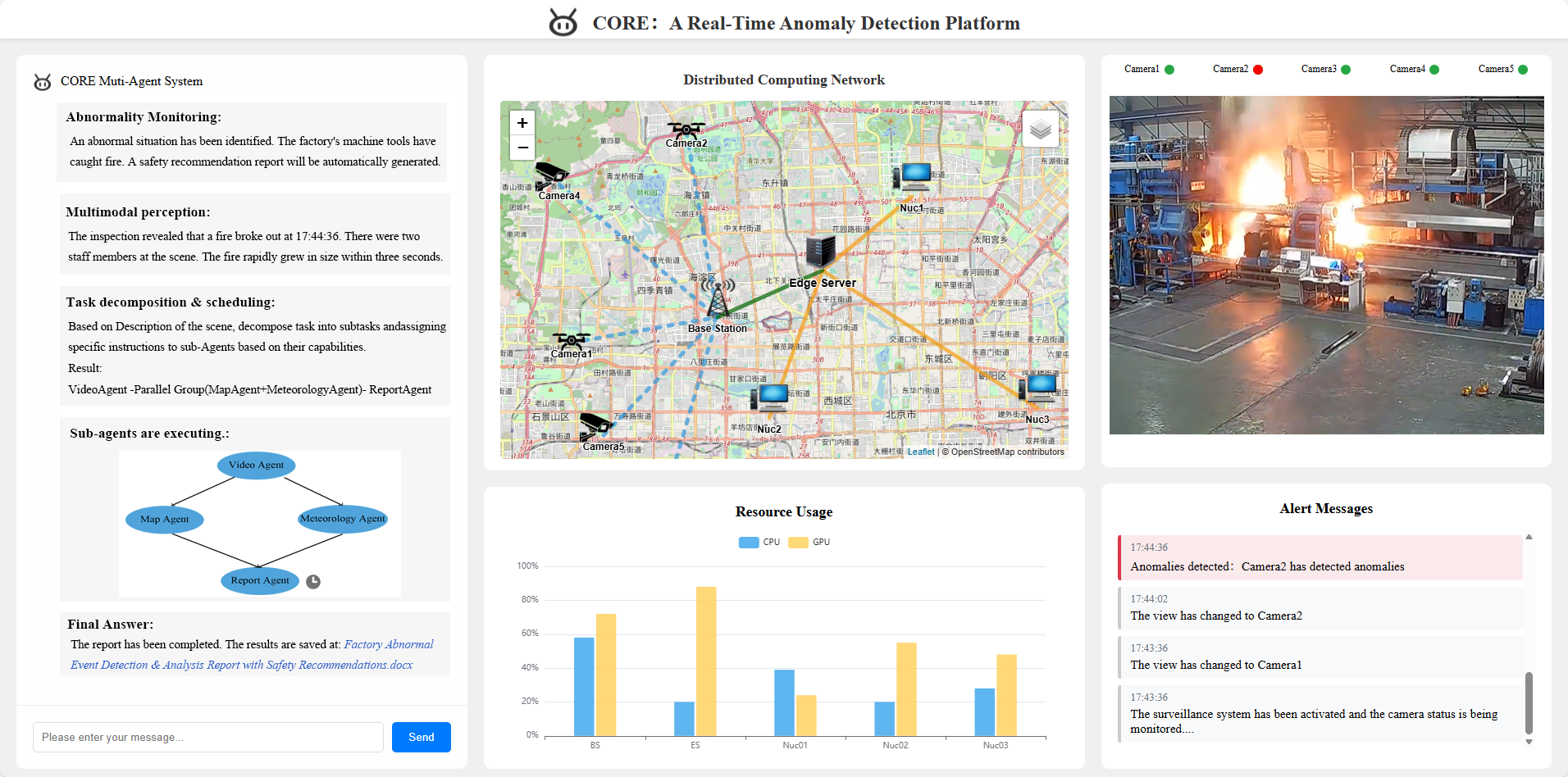}
    \caption{The CORE platform for real-time anomaly detection.}
    \label{fig:UI}
\end{figure}

The Collaborative Orchestration Role at Edge (CORE) framework was implemented on a real-world edge computing platform, illustrating its potential to transform time-sensitive applications, such as industrial automation and emergency response. As shown in Figure~\ref{fig:UI}, the user interface of the platform offers intuitive control and monitoring. A significant application of CORE is its use in the analysis of highway vehicle fire incidents (Figure~\ref{fig:algo_flow}), where CORE integrates multi-modal data, including video feeds, geographical information, and weather conditions, to develop comprehensive emergency response plans. CORE achieves high-accuracy and low-latency inference in real-time decision-making by integrating multilevel computing power networks, which is crucial for mitigating the impact of such accidents and enhancing emergency management.

\subsection{Experiment Deployment Environment}

The deployment environment was a hierarchical edge-cloud architecture tailored for low-latency and high-efficiency task processing purposes. At the core of this setup, the Deepseek-R1-distill-llama-70B model served as the task scheduler, running on eight NVIDIA A40 GPUs, each equipped with 48GB of VRAM. This configuration provides robust computational power for dynamic role orchestration and decision-making. At the edge, the MiniCPM-v2.6 model (8B parameters) was deployed on devices powered by NVIDIA 3090 and 4090 GPUs, enabling localized processing and reducing the dependency on cloud resources. Inter-agent coordination is facilitated by the Model Context Protocol (MCP), which ensures seamless data sharing and task execution across the system. This setup supports pipeline parallel execution and adaptive task allocation and optimizes performance under varying network conditions.

\subsection{Performance Evaluation}

\subsubsection{Metrics and Benchmarks}

The performance was evaluated using two key metrics: the task completion rate and latency. The task completion rate measures the percentage of successfully completed tasks across three difficulty levels: easy (0--3 tools), medium (4--6 tools), and hard (7--9 tools). Latency is divided into scheduling latency (task arrival to assignment) and execution latency (assignment to completion). These metrics were compared with single-agent algorithms (ReAct~\cite{Yao2022}, LLMCompiler~\cite{Kim2024}) and multi-agent algorithms (Static\_Dual\_Loop, Crew\_Ai~\cite{CrewAI}). Static\_Dual\_Loop is is a multi-agent manufacturing dual-loop architecture with fixed role arrangements. The evaluation dataset comprised an open-source video dataset (\url{https://github.com/steffensbola/furg-fire-dataset}) of anomalous behaviors (e.g., vehicle fires) and 300 task instructions generated by a large language model, manually curated for accuracy and representativeness.

\subsubsection{Results and Analysis}

\begin{figure}[t]
    \includegraphics[width=0.48\textwidth]{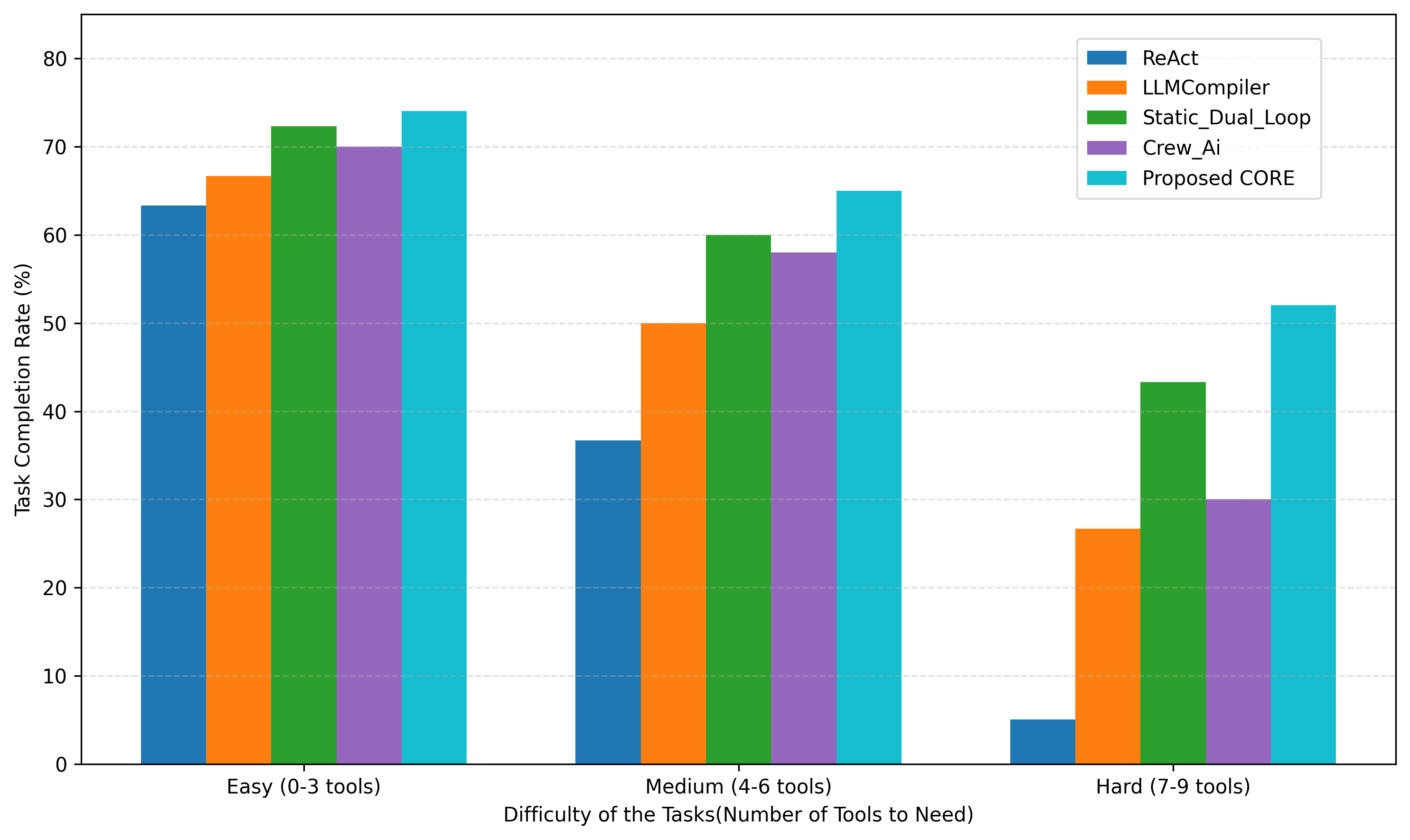}
    \caption{Task completion rate comparison.}
    \label{fig:success_rate}
\end{figure}

The experimental results indicate that CORE demonstrates superior task completion rates and reduced latency, particularly in complex scenarios (Figure~\ref{fig:success_rate}). CORE significantly outperforms single-agent methods, such as ReAct and LLMCompiler, across all levels of difficulty and exceeds the performance of Static\_Dual\_Loop by 25\% and 20\% on Medium and Hard tasks, respectively. This superiority is attributed to CORE's dynamic task decomposition and specialized multi-agent collaboration of CORE, which effectively address the scalability limitations inherent in single-agent systems and the inflexibility of static scheduling in managing multi-tool dependencies.

\begin{figure*}[t]
    \includegraphics[width=0.9\textwidth]{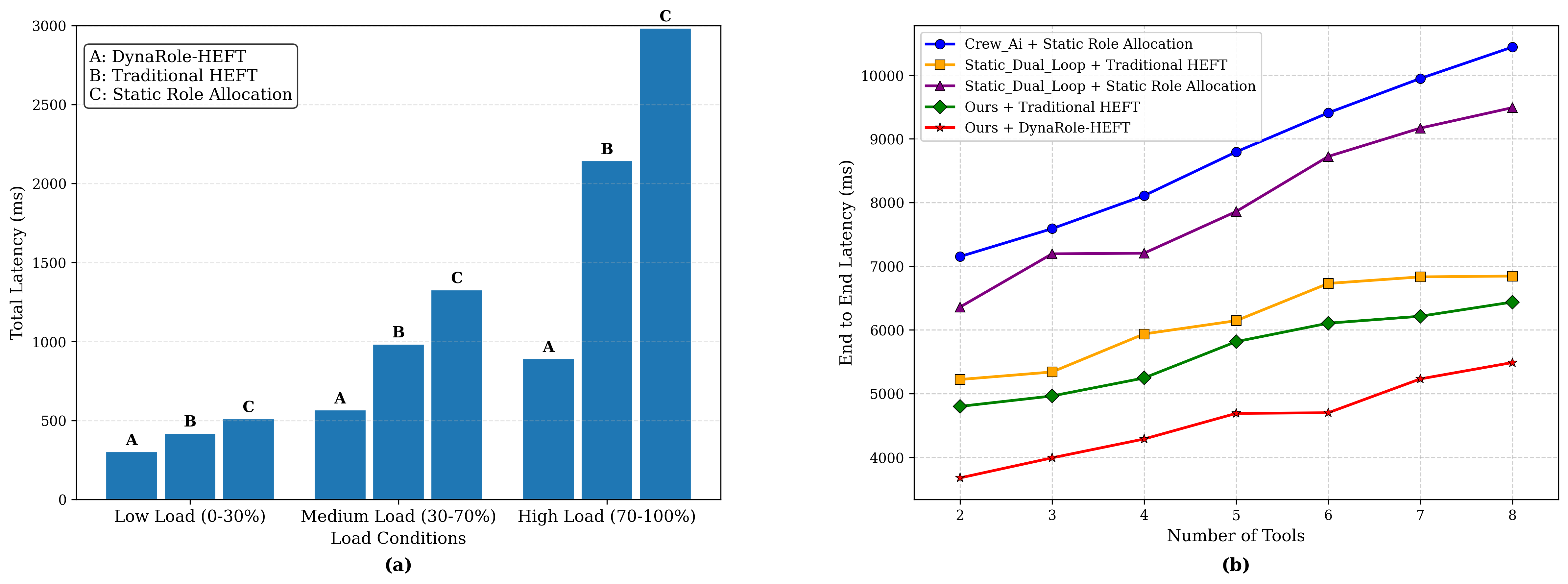}
    \centering
    \caption{Latency comparison: (a) Latency breakdown under varying loads; (b) End-to-end latency with increasing tool count.}
    \label{fig:latency}
\end{figure*}

The latency performance, as illustrated in Figure~\ref{fig:latency}, underscores the efficiency of the CORE. Subfigure (a) demonstrates that DynaRole-HEFT, the approach employed by CORE, surpasses the traditional HEFT and static assignment methods across various load conditions, exhibiting significantly reduced latencies and effective task allocation, which is advantageous for time-sensitive applications. Subfigure (b) depicts the scalability of CORE. Compared with other methodologies, CORE combined with DynaRole-HEFT maintains competitive end-to-end latency and significantly outperforms the static allocation of Crew Ai in multi-tool tasks.

\subsubsection{Summary}

The quantitative findings substantiate the superiority of CORE over the existing methodologies. Specifically, DynaRole-HEFT achieves a 52\% reduction in high-load latency relative to the traditional HEFT and enhances the medium-task success rate by 25\% compared to Static\_Dual\_Loop.  These improvements are driven by the pipeline parallel execution and context-aware task management of CORE, which collectively enhance the system efficiency and reliability. However, challenges in complex task performance and hardware generalizability remain, underscoring the need for further optimization to fully realize the potential of CORE in diverse 6G ecosystems in the future.

Figure~\ref{fig:latency}(a) further demonstrates that although DynaRole-HEFT substantially decreases the overall latency, the inference time of the orchestrator (ranging from 180 to 320 ms on a server equipped with eight NVIDIA A40 GPUs) accounts for 25–40\% of the total end-to-end delay under high-load conditions. This underscores the need for the development of lightweight scheduler designs for applications requiring sub-10 ms latency, such as remote surgery.

\section{Conclusions and Future Directions}

The CORE framework facilitates the efficient deployment of 6G technology by collaboratively orchestrating LLM agents across mobile devices and edge servers. This orchestration forms a "collective AI brain" that provides rapid and scalable AI services in domains such as smart cities, healthcare, industrial automation, and emergency response. Empirical evaluations demonstrate its superior performance in terms of both the task completion rate and latency. To address the remaining overhead associated with the orchestrator, future research should focus on developing lightweight scheduler designs through model distillation and specialization, particularly for applications requiring sub-10ms latency, such as remote surgeries. Additional research directions include integration with 6G network slicing for enhanced resource control, exploration of quantum-inspired algorithms for complex scheduling tasks, and standardization of inter-agent communication protocols to ensure seamless operation across heterogeneous 6G ecosystems.

\section{Biography Section}\vspace{-53pt}
\begin{IEEEbiographynophoto}{Zitong Yu}
Zitong Yu (yztong@bupt.edu.cn) is currently pursuing an M.S. degree at Beijing University of Posts and Telecommunications, China. His research interests include UAV swarm coordination, multi-agent systems, and large language models.
\end{IEEEbiographynophoto}
\vspace{-50pt} 
\begin{IEEEbiographynophoto}{Boquan Sun}
Boquan Sun (boquan\_sun@bupt.edu.cn) is currently pursuing an M.S. degree at Beijing University of Posts and Telecommunications, China. His research interests include edge computing, UAV navigation systems, and large language models.
\end{IEEEbiographynophoto}
\vspace{-50pt}
\begin{IEEEbiographynophoto}{Yang Li}
Yang Li (ly209991@bupt.edu.cn) is currently pursuing a Ph.D. degree at Beijing University of Posts and Telecommunications, China. His research interests include device-assisted mobile edge networks, computing offloading, and resource allocation.
\end{IEEEbiographynophoto}
\vspace{-50pt}
\begin{IEEEbiographynophoto}{Zheyan Qu}
Zheyan Qu (zheyanqu@bupt.edu.cn) is currently pursuing an M.S. degree at Beijing University of Posts and Telecommunications, China. His current research interests include edge intelligence, multi-agent systems, and generative AI.
\end{IEEEbiographynophoto}
\vspace{-50pt}
\begin{IEEEbiographynophoto}{Xing Zhang}
Xing Zhang (zhangx@ieee.org) is a full professor with the School of Information and Communications Engineering, Beijing University of Posts and Telecommunications, China. His research interests are mainly in 5G/6G networks, satellite communications, edge intelligence, and Internet of Things.
\end{IEEEbiographynophoto}

\end{document}